\documentclass[sigconf,natbib=true]{acmart}
\renewcommand\footnotetextcopyrightpermission[1]{}
\AtBeginDocument{%
  \providecommand\BibTeX{{%
    \normalfont B\kern-0.5em{\scshape i\kern-0.25em b}\kern-0.8em\TeX}}}

\setcopyright{acmcopyright}
\copyrightyear{2023}
\acmYear{2023}
\acmDOI{XXXXXXX.XXXXXXX}

\acmPrice{15.00}
\acmISBN{978-1-4503-XXXX-X/18/06}

\usepackage{multirow}
\usepackage{bbding}
\usepackage{bm}
\usepackage{colortbl}
\usepackage{xcolor}
\usepackage{subfigure}
\usepackage{arydshln}
\begin{document}

\title{Handling Label Uncertainty for Camera Incremental Person Re-Identification } 
\author{Zexian Yang$^{1,2}$ \quad Dayan Wu$^{1}$\footnotemark[1] \quad     Wanqian Zhang$^{1}$ \quad Bo Li$^{1,2}$ \quad Weiping Wang$^{1,2}$ \\
$^{1}$Institute of Information Engineering, Chinese Academy of Sciences \\ $^{2}$School of Cyber Security, University of Chinese Academy of Sciences\\
{\tt\small \{yangzexian,wudayan,zhangwanqian,libo,wangweiping\}@iie.ac.cn}}

\renewcommand{\shortauthors}{Yang, et al.}


\begin{abstract}
Incremental learning for person re-identification (ReID) aims to develop models that can be trained with a continuous data stream, which is a more practical setting for real-world applications. However, the existing incremental ReID methods make two strong assumptions that the cameras are fixed and the new-emerging data is class-disjoint from previous classes. This is unrealistic as previously observed pedestrians may re-appear and be captured again by new cameras. In this paper, we investigate person ReID in an unexplored scenario named Camera Incremental Person ReID (CIPR), which advances existing lifelong person ReID by taking into account the \textit{class overlap} issue. Specifically, new data collected from new cameras may probably contain an unknown proportion of identities seen before. This subsequently leads to the lack of cross-camera annotations for new data due to privacy concerns. To address these challenges, we propose a novel framework ExtendOVA. First, to handle the class overlap issue, we introduce an instance-wise seen-class identification module to discover previously seen identities at the instance level. Then, we propose a criterion for selecting confident ID-wise candidates and also devise an early learning regularization term to correct noise issues in pseudo labels. Furthermore, to compensate for the lack of previous data, we resort prototypical memory bank to create surrogate features, along with a cross-camera distillation loss to further retain the inter-camera relationship. The comprehensive experimental results on multiple benchmarks show that ExtendOVA significantly outperforms the state-of-the-arts with remarkable advantages.
 
\end{abstract}





\maketitle
\renewcommand{\thefootnote}{\fnsymbol{footnote}} 
\footnotetext[1]{Corresponding author.} 

\section{Introduction}
Person Re-IDentification (ReID) aims to match the same identity across non-overlapping camera views. The success of the modern offline supervised person ReID paradigm ~\cite{zheng2015scalable,song2019generalizable,xiao2016learning} is largely attributed to the availability of large-scale cross-camera annotations and the assumption that the surveillance system is fixed. The problem arises when the model needs to acquire new knowledge from newly installed cameras over time, which may require re-collecting data and retraining the model. However, manually establishing cross-camera annotations of all the identities from new and old cameras and then retraining them is expensive and cumbersome. Moreover, those methods are susceptible to catastrophic forgetting~\cite{mccloskey1989catastrophic} when adapted to real-world dynamic surveillance systems, particularly when data privacy concerns are taken into account.

\begin{figure}[t]
 \vspace{1.0cm}
  \centering
  \includegraphics[width=\linewidth]{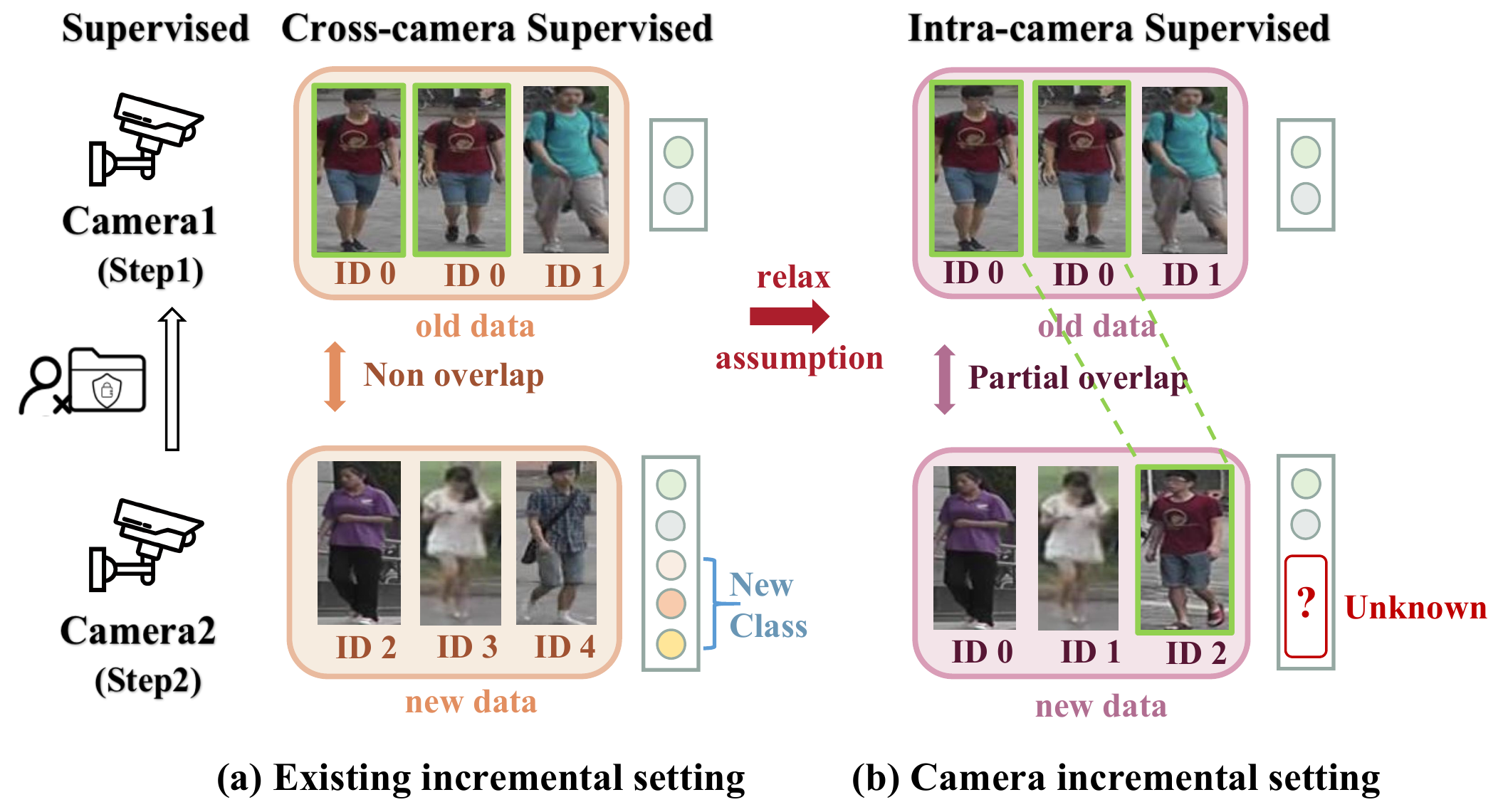}
  \caption{The comparison between camera incremental setting and
previous setting in incremental person ReID. (a) The existing setting assumes that identities in new data are completely disjoint with previous data. (b) Our setting relaxes the strict class-disjoint assumption. Under our camera incremental setting, the new data will only have intra-camera annotations, and may also contain previously seen people.}
  \label{fig:1}
 \vspace{0.5cm}
\end{figure}

Recently, there has been increasing attention ~\cite{wu2021generalising,huang2021lifelong,pu2021lifelong} on incremental learning (or lifelong) for person ReID (ILReID), which aims to address the practical requirements of continuously learning person ReID models from a stream of incoming data. As new data arrives, old data is not available for re-training due to privacy concerns. However, as shown in Fig.~\ref{fig:1}(a), existing ILReID methods commonly assume that the classes of new data are entirely different from the old ones. This assumption is not consistent with real-world scenarios, as previously observed pedestrians may \textit{\textbf{re-appear}} and be captured again by the camera.


Motivated by this gap, in this paper, we introduce a new task setting named Camera Incremental Person ReID (CIPR), that naturally meets the demand of incrementally updating the model from newly installed cameras without access to previous data. As shown in Fig.~\ref{fig:1}(b), unlike previous incremental setting in person ReID~\cite{wu2021generalising,huang2021lifelong,pu2021lifelong} that heavily relies on the class-disjoint assumption, the proposed CIPR allows for \textbf{partial class overlap between the old and new cameras}. In fact, previous methods have \emph{overlooked} a critical limitation, that annotations in person ReID are based solely on numerical IDs to distinguish between individuals, rather than specific categories (e.g. "cat"). This means that when previous data is no longer available, it becomes difficult to determine whether new data belongs to an existing or a new class, resulting in uncertainty in cross-camera labels of the new data. This thus makes CIPR a more realistic scenario since annotations can only be performed independently for each camera. Despite the above differences, CIPR still faces the risk of catastrophic forgetting due to the lack of prior data.
In general, the challenges of CIPR stem from two main aspects: 1) How to recognize and associate seen classes without any prior data (termed as class-overlap issue). As these seen classes should not be expected to learn as new ones, any accumulated errors can lead to performance degradation over time. 2) How to learn more informative knowledge from new cameras while also retaining previously acquired knowledge.

To handle the above challenges in CIPR, a novel framework ExtendOVA is proposed. Specifically, to eliminate the detrimental effect of the class overlap issue, we first incorporate an One-vs-All (OVA) detector~\cite{padhy2020revisiting} that can identify unknown samples from new data. Nevertheless, directly applying the vanilla OVA detector to the CIPR task is problematic for two main reasons. 
On one hand, the OVA detector only models instance-level recognition, which fails to inherently identify whether a given class is unseen or not. On the other hand, the OVA detector is trained on the original camera data, leading to the domain shift from the new camera. As a result, potentially seen classes will be misidentified as unseen classes. To achieve ID-wise cross-camera identification, we extend the OVA detector by 1) We propose a simple yet effective criterion for selecting confidence-seen classes. 2) We devise an early learning regularization term to address concerns of domain shift and rectify potential noisy labels. In addition, to compensate for the lack of previous data against the second challenge of CIPR, we resort to the prototypical memory bank to create surrogate features based on the prototypes and the Batch-Normalization (BN) layer statistics. We also present a cross-camera distillation loss to retain the inter-camera relationship. In conclusion, our contributions can be summarized as follows:
\begin{itemize}
\item  We introduce a novel yet more practical ReID task, named \emph{Camera Incremental Person ReID} (CIPR), which is fundamentally different from the existing lifelong person ReID tasks. It demands continuous learning of more generalizable representations through data from newly installed cameras only with intra-camera supervision.


\item
We carefully design a novel framework ExtendOVA, which crafts an ID-wise pseudo label generation module against the peculiar class overlap issue under the camera incremental setting.

\item  For extensive assessment, we build a simple baseline in addition to ExtendOVA to tackle CIPR.  Experimental results show that the proposed approach gains significant advantages over the comparative methods.
\end{itemize}

\begin{figure*}[t]
  \centering
  
  \includegraphics[width=0.95\linewidth]{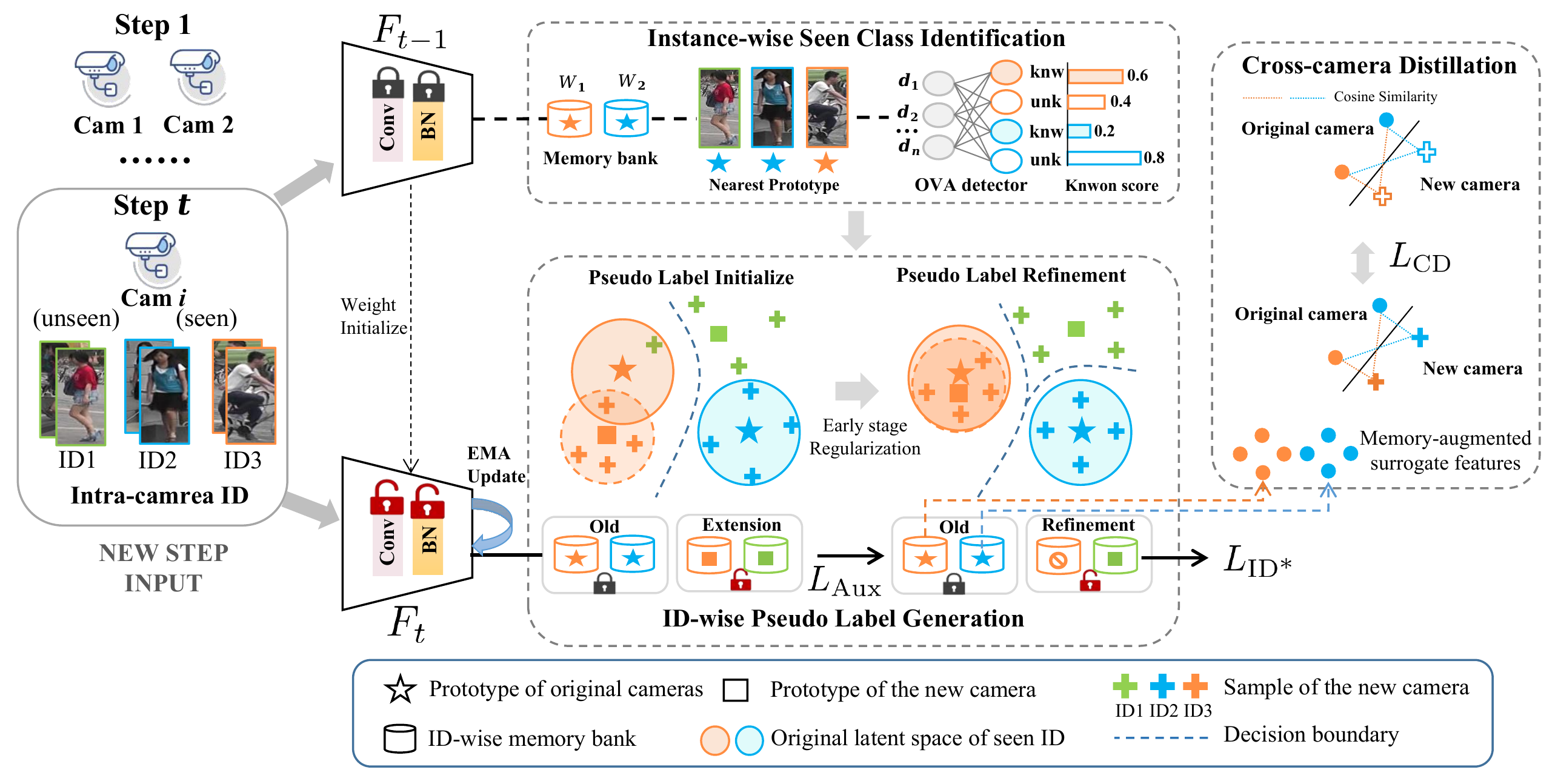}
  \caption{The proposed framework consists of three parts. The first part called Instance-wise Seen Class Identification, is used for detecting the seen and unseen samples using a One-vs-All detector before training. The second part is to generate ID-wise pseudo labels and further correct noisy labels by $\mathcal{L}_\text{Aux}$ at the early training stage. The third part is cross-camera distillation, which leverages sampled surrogate features to regularize forgetting by forcing relationship to be maintained between cameras.}
  \label{fig:framework}
\end{figure*}

\section{Related Work}
\label{sec:2}

\noindent{\textbf{Person Re-identification.}}
Offline person ReID settings can be roughly distinguished into three categories: supervised person ReID, unsupervised person ReID, and intra-camera supervised person ReID. Supervised person ReID ~\cite{luo2019alignedreid++,sun2018beyond,wang2020high} are usually superior in performance but are less scalable, relying on a large amount of cross-camera annotations. Differently, unsupervised person ReID~\cite{yu2017cross,yu2018unsupervised,wu2020tracklet,fu2019self,zou2020joint} is more challenging employing either clustering algorithms to generate pseudo labels or the extra source
labeled data to boost the performance. Moreover, intra-camera supervised (ICS) person ReID is another perspective to reduce annotation labor ~\cite{ge2021cross,zhu2019intra,peng2022consistent}, where cross-camera association labels are removed from the training data. However, these settings assume that the training data is pre-collected, and thus camera relations can be learned by matching cross-camera images, yet they are not suited to incrementally add new cameras over time. Our task of incrementally learning person ReID models from newly installed cameras is related to the problem of intra-camera supervision, but with the added challenge of privacy concerns related to cross-camera images for associating positive pairs.

\textbf{Lifelong Person Re-identification.} Recently, there has been significant interest in incremental learning for Person ReID, which aims to continuously learn new knowledge without experiencing catastrophic forgetting ~\cite{ratcliff1990connectionist}. Various methods have been proposed to prevent forgetting, which can be broadly divided into two categories: replay-based and data-free methods. Replay-based methods rely on maintaining a memory bank of limited samples that are recorded for replay. However, this approach requires additional storage, and maintaining raw data poses a risk to privacy. In contrast, data-free methods do not rely on any old samples. In this paper, we focus on a data-free incremental learning pipeline. 

Previous research ~\cite{pu2021lifelong,wu2021generalising,lu2022augmented,DBLP:conf/aaai/GeDWXYHZ22} primarily focuses on incremental scenarios, where new identities
keep increasing in fixed camera systems. However, contemporary
surveillance systems are dynamic, meaning cameras can be installed at any time. In this paper, we consider a more practical scenario for lifelong person re-id, which aims to optimize the model when one or more cameras are introduced in the existing surveillance systems. Our approach does not require any strict class-disjoint assumption for model training, and it also considers a scenario where cross-camera labels are unavailable in training data. 

\textbf{Out-of-Distribution Detection.}
Out-of-distribution (OOD) detection is a binary classification problem that involves the ability of a model to distinguish between in-distribution and out-of-distribution samples during inference. There are various approaches to OOD detection, some of which involve modeling different scoring functions, such as maximum softmax probability~\cite{hendrycks2016baseline,DBLP:conf/iclr/LiangLS18} or entropy~\cite{liu2020energy,chan2021entropy}, to estimate confidence and identify OOD samples. Others~\cite{zong2018deep,pidhorskyi2018generative}  utilize generative models to learn the distribution of in-distribution data.  One approach~\cite{padhy2020revisiting} proposed in a recent paper involves using neural One-vs-All (OVA) classifiers to handle out-of-distribution detection. In our work, we incorporate the OVA detector to differentiate between "unseen" and potential "seen" samples. However, it's important to note that the OVA detector is unable to perform ID-wise prediction and may not be robust enough to handle data with domain gaps.

\section{Preliminary}
\label{sec:3}

\subsection{Problem Formulation}
Consider a CIPR problem with several steps, and each incremental step introduces a new camera with a set of classes to learn. Formally, in the $t$-th step, we have the training data $\mathcal{D}_t=\{\bm{\mathit{X_t,Y_t}}\}=\{(x^t_i,y^t_i)|^{N_t}_{i=1} \}$ with intra-camera annotations $y^t_i\in \bm{Y_t}$ captured by the newly installed camera $c_t$, and $N_t$ is the number of classes in $c_t$.  We note that the training data $\mathcal{D}_{t}$ can contain overlapping classes in $\mathcal{D}_{t-1}$, while the old training data $\mathcal{D}_{t-1}$ are not available due to the privacy concern. Hence, we first need to identify the real number of extensions to correctly learn new classes. The goal of CIPR is to learn a robust ReID model that can be generalized to unseen classes from all encountered cameras.

\subsection{A CIPR Baseline}
\label{sec:baseline}
We first present a straightforward baseline for CIPR task. Basically, in the $t$-th step ($t>1$), the feature extractor $F(\theta_{t})$ initialized by $F(\theta_{t-1})$ is updated to learn a set of classes employing $\mathcal{D}_{t}$, and the classifier $G(\phi_{t})$ is also extended to the corresponding new dimension~\cite{hou2019learning}, which is expected to predict all the classes seen so far. As a common incremental learning baseline, in addition to ReID loss~\cite{he2020fastreid} (e.g. ID loss $\mathcal{L}_\text{ID}$+ triplet loss $\mathcal{L}_\text{Triplet}$~\cite{hermans2017defense}), knowledge distillation (KD) loss  $\mathcal{L}_\text{KD}$ is employed to prevent catastrophic forgetting, which can be formulated as: 
\begin{equation}
\label{eq:kd}
    \mathcal{L}_{\text{KD}} = \sum_{i\in \bm{X_t}}KL(p^n_i||p^o_i),
\end{equation}
where $KL(\cdot)$ is the Kullback Leibler (KL) divergence,  $p^o_i$ and $p^n_i$ denote the logit output of the old and new models, respectively.

To discriminate the seen and unseen identities without accessing the old data, a straightforward method is to leverage the softmax prediction score. We assume that samples belonging to unseen classes will produce smooth probability distributions since they are equally wrong and ambiguous. Therefore, we can treat an image as the seen class if the maximum softmax score is above a threshold $T$. For samples identified as a new class, we add a new ID based on the existing old classes. For samples classified into old classes, we use the model predict as its pseudo label. Then we can minimize the cross entropy with the global pseudo labels. The loss function can be formulated as: 

\begin{equation}
      \mathcal{L}_{\text{ID}}  =  \mathcal{L}_{\text{CE}}(G(F(\bm{X_t};\theta_{t});\phi_{t}),\bm{Y'_t}).
\end{equation}
where $\bm{Y'_t}$ is the pseudo label of samples $\bm{X_t}$, $\mathcal{L}_{CE} $ is the cross-entropy loss function.

Overall, the optimization objective of the baseline CIPR model can be formulated as:

\begin{equation}
    \mathcal{L}_\text{Base} =  \mathcal{L}_{\text{ID}} +  \mathcal{L}_{\text{Triplet}} + \lambda_0 \mathcal{L}_{\text{KD}}.
\end{equation}

\section{METHODOLOGY }
\label{sec:4}
The filtering mechanism proposed in our baseline method is an alternative way to address the class-overlap issue. However, the manual set threshold $T$ is not robust enough to identify old classes, and the classifier is biased toward mass classes over few classes~\cite{hendrycks2016baseline}. Therefore, in this section, we introduce a new framework for CIPR.

\subsection{Overview of Framework}
\label{sec:overview}
The graphical illustration of our framework is depicted in Fig. \ref{fig:framework}. We replace the linear classifier with the non-parametric memory bank to alleviate the over-confidence issue ~\cite{bendale2016towards}, which stores the moving average of the cluster prototypes. The old model is fixed, and the new model is updataed via Exponential Moving Average (EMA) scheme during the optimization. Then we elaborate our ExtendOVA in three parts to tackle CIPR problem. The first technical novelty comes from taking advantage of the One-vs-All (OVA) detector for instance-wise seen class identification before training (section~\ref{sec:ISCI}). Then the samples are assigned with global pseudo labels via our criteria and an early learning regularization term $\mathcal{L}_{\text{Aux}}$, as to be detailed in section~\ref{sec:IPLG}.
Finally, in section~\ref{sec:CD}, surrogate features are sampled based on the  memory bank to guide the cross-camera distillation objective.

\subsection{Instance-wise Seen Class Identification }

\label{sec:ISCI}

In this section, we elaborate on the process of instance-wise seen class identification. We first describe the training of the One-vs-All detector before describing the remaining methods.

\noindent{\textbf{One-vs-All Detector.}} The One-vs-All (OVA) detector~\cite{padhy2020revisiting,saito2021ovanet} is first proposed for the out-of-distribution detection, which extends a binary classifier to a multi-class classifier to learn a boundary between in-liers and outliers.  Specifically, the OVA detector consists of multiple binary sub-classifiers, each of which is trained to distinguish that class from all other classes, i.e., samples belonging to this class are positive while others are negative. For more effectively learning a boundary to identify unknown identities, herein we only pick hard negative samples to compute the loss. Formally, we denote $p(\hat{y}^c|x)$ as the positive softmax output for the class $c$. The optimization objective for a sample $x_i$ within label $y_i$ can be formulated as:

\begin{equation}
    \mathcal{L}_{\text{ova}}(x_i,y_i) = -\log{p(\hat{y}^{y_i}|x_i)}-\min_{c\neq y_i}\log{(1-p(\hat{y}^c|x_i))}.
    \label{eq1}
\end{equation}

\noindent{\textbf{Seen Class Identification.}} When new data arrives, we first get the nearest  prototype and take the corresponding sub-classifier output of the OVA detector to determine whether the sample is a seen or unseen class, as illustrated in Fig. \ref{fig:framework}. Essentially, each sub-classifier of the OVA detector corresponds to the latent space of its class, and if a sample exceeds all the boundaries of that space, it will be recognized as outliers. The advantage of the OVA detector is that it can learn an adaptive threshold between seen and unseen classes.



\subsection{ID-wise Pseudo Label Generation}
\label{sec:IPLG}

Although the OVA detector is effective and more robust for instance-level prediction, it may still introduce noisy labels, especially for hard samples. As a result, two images of the same class may be paradoxically predicted as a new class and an old one, which can affect the class-level prediction. Furthermore, we argue that due to the domain gap, the latent space trained for each class could not effectively represent the data from future cameras, as illustrated in Fig.~\ref{fig:framework}. To this end, we propose an ID-wise pseudo label generation module (IPLG) to correct noisy labels and associate the samples with the same local label to the same pseudo-global label. We shall detail the operation of this module below.

\noindent{\textbf{Pseudo Label Initialize.}} We propose a simple criterion to select confident-seen classes. Given a batch of samples $\{(x_i^t,y_i^t)\}^{B}_{i=1}$ that follows PK sampling, we first analyze the output of the samples with the same label $y_i^t$ from the OVA detector. An identity $y_i^t$ is the seen class if and only if all samples with label $y_i^t$ are predicted to belong to the seen class. Denoting the set of seen classes as $C_{sc}$, for $y_i^t \in C_{sc} $, we use the nearest class predicted by the prototype classifier as pseudo labels. For the remaining classes (denoted by $C_{uc}$) that are excluded by our criterion, we re-label them to a new class. Formally, the pseudo labels are assigned in the following way:

\begin{equation}
  \hat{y}_{i}=
\begin{cases}
\underset{k}{\arg \max } ~{W}_{k}^{\top} F(x_i^t;\theta_t), & y_i^t \in C_{sc}, k \in [1,N_{t-1}] \\
N_{t-1}+n ~, & y_i^t \in C_{uc}, n\in [1,|C_{uc}|],
\end{cases}
 \label{eq:5}
\end{equation}
where $W_k \in \mathcal{R}^d$ stands for the $k$-th column of ID-wise prototype in memory bank $\bm{W}\in \mathcal{R}^{d\times N_{t-1}}$, $d$ is the feature dimension. Note that we only choose the most frequently predicted one if there are multiple $k$ for one class. Correspondingly, we expand the memory bank to $\bm{W}\in \mathcal{R}^{d\times N_t' }$, $N_t'= N_{t-1}+|C_{uc}|$.

\noindent{\textbf{Pseudo Label Refinement.}} To rectify the noisy pseudo labels caused by the domain shift, an auxiliary loss is designed to regularize the early-stage learning.  Concretely, leveraging the initialized pseudo labels, the network can be trained via a softmax loss to identify both seen and unseen classes in the new data, ensuring that features are closest to their corresponding prototypes, which can be formulated as:

\begin{equation}
    \mathcal{L}_{\text{ID}^*} = -\sum_{x^t_i \in \mathcal{D}_t} \log \frac{e^{W_{\hat{y}_i}^{\top}F(x_i^t;\theta_t)} }{e^{W_{\hat{y}_i}^{\top}F(x_i^t;\theta_t)}+ {\textstyle \sum_{k=1,k\neq \hat{y}_i}^{N'}}e^{W_{k}^{\top}F(x_i^t;\theta_t)} },
    \label{eq:6}
\end{equation}
the key is to keep the old prototypes fixed throughout the process, serving as a boilerplate for learning domain-invariant features. 

Additionally, the hope is that samples belonging to unseen classes can also output the second largest probability to the potential nearest prototype, which is constrained by the following auxiliary loss:

\begin{equation}
    \mathcal{L}_{\text{Aux}} = -\sum_{\hat{y}_i>N_{t-1}} \log \frac{e^{W_{\tilde{y} _i}^{\top}F(x_i^t;\theta_t)} }{e^{W_{\tilde{y}_i}^{\top}F(x_i^t;\theta_t)}+ {\textstyle \sum_{k=1,k\neq \tilde{y} _i}^{N_{t-1}}}e^{W_{k}^{\top}F(x_i^t;\theta_t)} }. 
\end{equation}
Here, $\tilde{y}_i\in[1,N_{t-1}]$ is obtained using the same way as in Eq.\ref{eq:5}.  The motivation behind this approach lies in the consensus ~\cite{ren2018meta,grandvalet2004semi} that seen classes are usually clustered to form high-density regions in the latent space. Hence, this regularizer encourages these samples to be closed to a shared and real prototype.  On the other hand, the unseen classes are often distributed in low-density regions, leading to optimization conflicts where the samples struggle to simultaneously approach the current prototype and the nearest old prototype. 

After the early-stage regularization, we would employ the selection criterion again to obtain refined pseudo labels, which would continue to be used for model updates. Meanwhile, those new prototypes that were previously created by falsely identifying as unseen classes would be removed from the memory bank.


\subsection{Cross-camera Distillation}
\label{sec:CD}


 In incremental learning, models are susceptible to forgetting previous learned knowledge without relearning the old data. To address this, previous work~\cite{lu2022augmented} has proposed to use a GAN~\cite{goodfellow2020generative} to reconstruct old data in the image space. In our method, we generate substitute samples in the feature space instead. Concretely, to estimate the distribution of the old data, we assume a class-conditioned multivariate Gaussian distribution denoted as $F(x_i^{t-1}|y_i^{t-1}=k)\sim \mathcal{N}_k^{t-1}(\mu_k^{t-1}, {\textstyle \sum_{k}^{t-1} } ) $. Here, $\mu_k^{t-1}$ is the mean of the Gaussian distribution and can be approximated using our prototypical memory bank. To estimate the covariance matrices, we utilize the statistics of BatchNorm (BN) layers. During training, a BN layer normalizes the features, which implicitly captures the means and variances of the data~\cite{yin2020dreaming}, thereby enabling the estimation of the covariance matrices of the old data. Overall, we estimate the distribution of the data in previous step by:
\begin{equation}
\begin{aligned}
    \mu_k^{t-1} \simeq W_k  &&&
     {\textstyle \sum_{k}^{t-1} } \simeq \text{BN}(var)
\end{aligned}
\end{equation}

Then we can sample surrogate features $\tilde{f_k} \sim \tilde{\mathcal{N}}_k(W_k , \text{BN}(var) )$. Based on these surrogate features, we present a cross-camera distillation loss that serves to regularize forgetting, by ensuring that the cosine distance is maintained across different cameras. Formally, given a batch of samples $\bm{\mathcal{X}}=\{(x^t_i,\hat{y}_i)|^{B}_{i=1} \}$ along with a batch of sampled surrogate features $\bm{\tilde{F}} = \{\tilde{f_{k_1}},\tilde{f_{k_2}}...\tilde{f_{k_B}} \}$, the loss can be calculated by:

\begin{equation}
\begin{aligned}
\mathcal{L}_\text{CD} =  \left \| \cos(F(\bm{\mathcal{X}};\theta_t),\bm{\tilde{F}})-\cos(F(\bm{\mathcal{X}};\theta_{t-1}),\bm{\tilde{F}}) \right \|_2^2.    
\end{aligned}
\end{equation}

where $\cos (a,b)=\frac{a\cdot b}{\left \| a \right \|_2 \left \| b \right \|_2 } $ denotes the cosine similarity . The distillation loss $\mathcal{L}_\text{CD}$ improves stability that is commensurate with the ability of previous data to maintain past structure.

\subsection{Optimization Summary}

In summary, the overall objective function for our ExtendOVA framework is formulated as

\begin{equation}
    \mathcal{L} =  \mathcal{L}_{\text{Triplet}}+ \mathcal{L}_{\text{ID}^*} + \lambda_1 \mathcal{L}_{\text{Aux}} + \lambda_2\mathcal{L}_\text{CD}.
\label{eq:10}
\end{equation}
where $\lambda_1$ and $\lambda_2$ are coefficients. To enhance the model's stability during optimization, we utilize the Exponential Moving Average (EMA) technique~\cite{tarvainen2017mean}, wherein the student model's parameters are initially shared by the teacher model $F(\theta_{t})$.  Once this iteration is complete, the student model is updated using the EMA parameters computed from the teacher model's parameters by
\begin{equation}
    \theta_{s,t} = \alpha \theta _t +  (1-\alpha )\theta _{s,t-1}
\end{equation}
where $\alpha$ is a smoothing factor typically set to a value 0.99~\cite{xu2021end}. In the test phase, we will use the student model to extract feature representations. By incorporating EMA into the training process, the updates to the student model's parameters are smoothed, leading to improved stability and better generalization performance.

\section{Experiments}
\label{sec:5}
\subsection{Datasets and Evaluation Metrics}
\begin{figure}[t]
  \centering
 \setlength{\abovecaptionskip}{-0.01cm}
  \includegraphics[width=\linewidth]{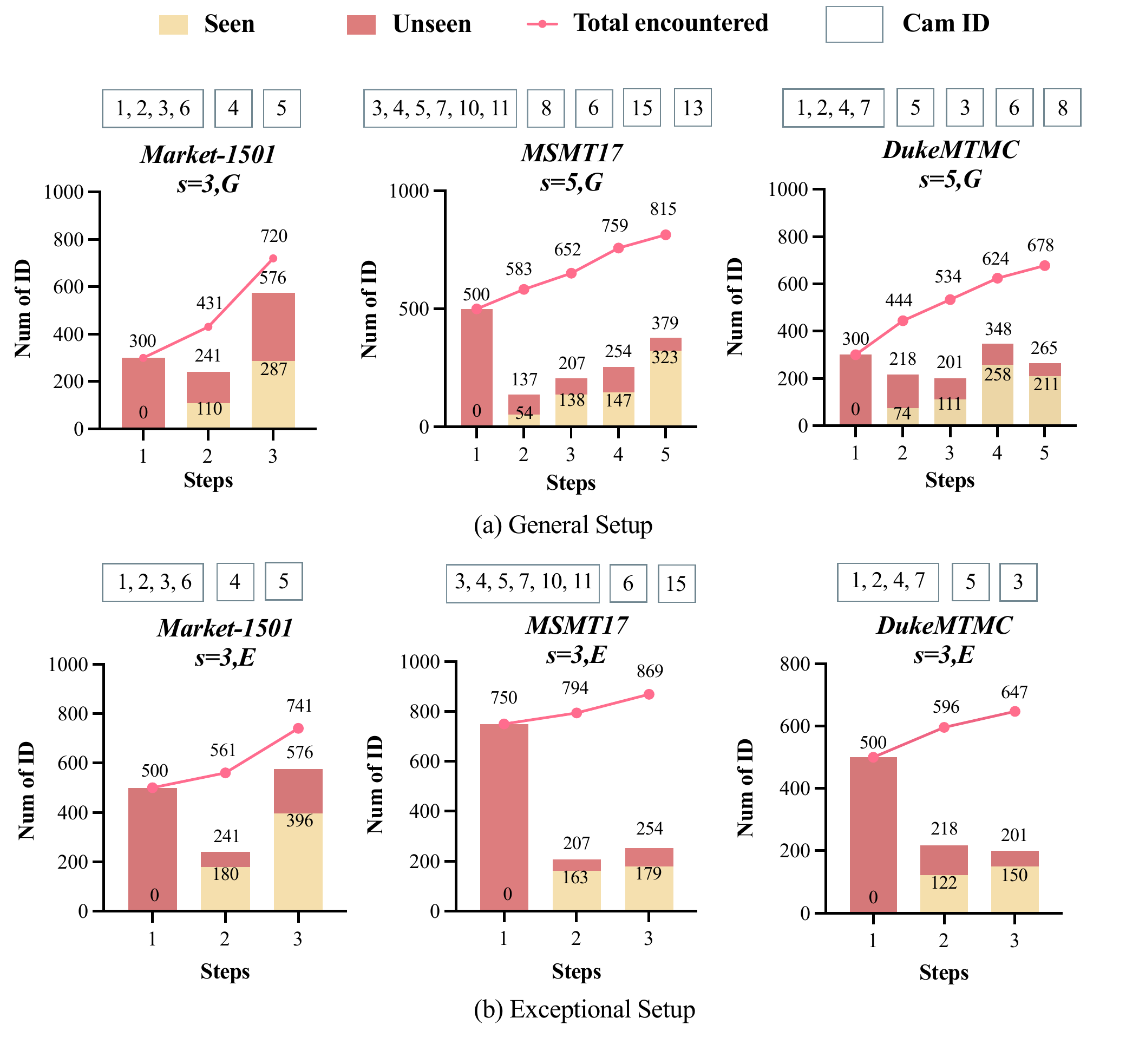}
  \caption{The distribution of identity number throughout training with different setups.}
  \label{fig:3}
\end{figure}

\noindent{\textbf{Datasets.}}  To evaluate and compare different methods
under Camera Incremental Person Re-Identification (CIPR) setting, three large-scale person Re-Identification (ReID) datasets Market-1501~\cite{zheng2015scalable}, MSMT17~\cite{wei2018person} and DukeMTMC~\cite{zheng2017unlabeled} (only for academic use, without displaying images of persons) are exploited.  We form the intra-camera annotations based on the provided labels. In order to simulate a realistic scenario for incremental learning in person re-identification, we simulate the deployment of a surveillance system that starts with multiple cameras and gradually adds new cameras over time. For example, we select 4 cameras from the Market-1501 for initial training and incrementally add 1 more camera in each subsequent step. Similarly, we create a five-step incremental training setup for MSMT17 and DukeMTMC. It is worth noting that we do not employ all classes of the initial cameras in the first step, but instead perform sampling to generate multiple setups to suit various conditions. The statistics of the
datasets are shown in Fig.~\ref{fig:3}; section~\ref{sec:implementation} goes into detail about the setups. 

\noindent{\textbf{Testing Protocols.}} Two commonly used metrics mean Average Precision (mAP) and Rank-1(R-1) accuracy are used to evaluate the performance of CIPR. To measure the model's ability to adapt and learn new knowledge, we evaluate its performance on unseen classes of all encountered cameras during the incremental learning process. Additionally, to assess the model's anti-forgetting ability, we evaluate its performance on unseen classes of the initial cameras as a measure of retention of previously learned knowledge.

\begin{table*}[t]
\centering
\caption{Comparison of the final-step incremental results with the state-of-the-art methods in different setups. Joint-T refers to the upper-bound result. {\color[HTML]{FE0000}Red} and {\color[HTML]{3531FF}blue}: the best and second-best results.}

\begin{tabular}{lc|cccc|cccc|cccc}
\toprule[1pt]
\multicolumn{1}{l}{}                          & \multicolumn{1}{c|}{}                            & \multicolumn{4}{c|}{Market-1501}                                                                                                            & \multicolumn{4}{c|}{MSMT17}                                                                                          & \multicolumn{4}{c}{DukeMTMC}       \\
\multicolumn{1}{l}{}                          & \multicolumn{1}{c|}{}                            & \multicolumn{2}{c}{\textit{s=3,G}}                        & \multicolumn{2}{c|}{\textit{s=3,E}}                                             & \multicolumn{2}{c}{\textit{s=5,G}}                        & \multicolumn{2}{c|}{\textit{s=3,E}}                      & \multicolumn{2}{c}{\textit{s=5,G}}&\multicolumn{2}{c}{\textit{s=3,E}} \\ \cmidrule(lr){3-4} \cmidrule(lr){5-6} \cmidrule(lr){7-8}  \cmidrule(lr){9-10}  \cmidrule(lr){11-12}  \cmidrule(lr){13-14}  
\multicolumn{1}{l}{\multirow{-3}{*}{Method}} & \multicolumn{1}{c|}{\multirow{-3}{*}{Reference}} & mAP                         & R-1                      & mAP                         & R-1                                           & mAP                         & R-1                      & mAP                         & R-1                    & mAP             & R-1   & mAP             & R-1          \\ \hline
\multicolumn{12}{l}{\textit{Upper Bound and Lower Bound}}                                                                                                                                                                                                                                                                                                                                               \\ \hdashline
Joint-T                                       & -                                               & 73.5                        & 88.2                        & 79.1                        & \multicolumn{1}{c|}{88.4}                        & 45.4                        & 70.9                        & 52.1                        & \multicolumn{1}{c|}{75.2} & 69.1            & 81.8      &73.7&84.2       \\
Finetune                                      & -                                               & 40.1                        & 62.5                        & 45.5                        & \multicolumn{1}{c|}{67.4}                        & 20.3                        & 30.2                        & 21.7                        & \multicolumn{1}{c|}{30.5} & 32.7            & 48.3& 34.1& 46.3             \\ \hline
\multicolumn{12}{l}{\textit{Data-free methods}}                                                                                                                                                                                                                                                                                                                                                         \\ \hdashline
LwF~\cite{li2017learning}                                           & TPAMI'17                                         & 57.8                        & 78.5                        & 62.7                        & \multicolumn{1}{c|}{83.3}                        & 27.6                        & 60.4                        & 30.2                        & \multicolumn{1}{c|}{59.8} & 45.4            & 62.5    &51.3&68.3         \\
MMD~\cite{long2015learning}                                           & ICML'15                                          & 60.7                        & 79.7                        & 65.8                        & \multicolumn{1}{c|}{83.7}                        &   24.9                          &       58.6                      & 28.8                        & \multicolumn{1}{c|}{53.5} & 38.7            & 57.9      &51.4
&68.8       \\
AKA ~\cite{pu2021lifelong}                                          & CVPR'21                                          & 60.2                        & 70.3                        & 61.5                        & \multicolumn{1}{c|}{81.1}                        & 29.7                        & 57.9                        & 33.0                        & \multicolumn{1}{c|}{60.2} &   41.9          &      57.6  & 52.6 & 61.9       \\
AGD ~\cite{lu2022augmented}                                         & CVPR'22                                          & 54.4                        & 68.1                        & 58.6                        & \multicolumn{1}{c|}{76.1}                        & 30.5                        & 64.8                        & 36.8                        & \multicolumn{1}{c|}{62.1} &            46.0     &          65.6    &48.2
&64.9    \\ 

PatchKD~\cite{sun2022patch} & MM'22 &64.8 &82.5 & \color[HTML]{3531FF}71.2 &83.4 &27.9 & 45.2 &33.9 &54.2 &51.1 &67.8&51.3 & 70.1\\\hline
\multicolumn{12}{l}{\textit{Replay-based methods}}                                                                                                                                                                                                                                                                                                                                                      \\ \hdashline
iCaRL~\cite{rebuffi2017icarl}                             & CVPR'17                                          & 59.4                        & 82.0                        & 64.8                        & \multicolumn{1}{c|}{84.0}                        & {\color[HTML]{3531FF} 33.2} & {\color[HTML]{3531FF} 68.6} & {\color[HTML]{3531FF} 42.2} & \multicolumn{1}{c|}{69.1} &         50.7        &     68.5      &53.5 &70.8       \\
PTKP~\cite{DBLP:conf/aaai/GeDWXYHZ22}                                          & AAAI'22                                          & {\color[HTML]{3531FF} 68.1} & {\color[HTML]{3531FF} 84.8} &  69.3 & \multicolumn{1}{c|}{84.3}                        & 33.0                        & 67.9                        & 41.3                        & \multicolumn{1}{c|}{\color[HTML]{3531FF}69.4} &     \color[HTML]{3531FF} 52.3           &       \color[HTML]{3531FF} 68.9  &\color[HTML]{3531FF} 56.3&\color[HTML]{3531FF} 71.1         \\ \hline
\multicolumn{12}{l}{\textit{Our methods (Data-free)}}                                                                                                                            \\ \hdashline
Baseline                                      &                                                 & 61.7                        & 80.3                        & 66.9                        & \multicolumn{1}{c|}{{\color[HTML]{3531FF}85.0}} & 30.9                        & 67.0                        & 36.6                        & \multicolumn{1}{c|}{66.1} & 47.8            & 65.1   &  53.4 & 69.1        \\
ExtendOVA                      &                      &  \color[HTML]{FE0000} 70.3                     &  \color[HTML]{FE0000}86.6                          & \color[HTML]{FE0000}75.6                       & \multicolumn{1}{c|}{\color[HTML]{FE0000} 88.2}                     & \color[HTML]{FE0000}  35.3            &  \color[HTML]{FE0000}   70.2                          &  \color[HTML]{FE0000} 46.3                          & \multicolumn{1}{c|}{\color[HTML]{FE0000}74.4}     &  \color[HTML]{FE0000} 53.4               &     \color[HTML]{FE0000} 70.2      &  \color[HTML]{FE0000}62.4&  \color[HTML]{FE0000}75.8      \\ \bottomrule[1pt]
\end{tabular}
\label{tab:1}
\end{table*}

 \begin{figure*}[ht]
  \centering
   \setlength{\abovecaptionskip}{-0.01cm}
  \includegraphics[width=0.85\textwidth]{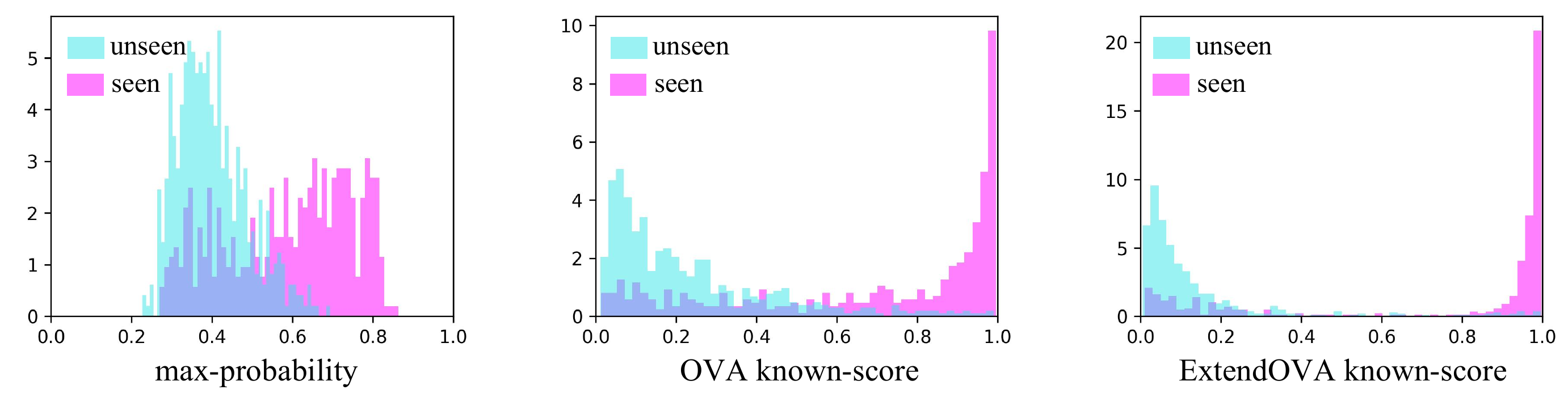}
  \caption{The confidence score distribution of seen and unseen samples produced by baseline (left), OVA detector (middle) and ExtendOVA (right) on Market-1501 in general setup.}
   \label{fig:4}
\end{figure*}

\subsection{Implementation Details}
\label{sec:implementation}
 We use the widely adopted ResNet-50~\cite{he2016deep} as the backbone network. To obtain 2048-dimensional features, a Batch Normalization (BN) layer ~\cite{ioffe2015batch} is placed after the last layer of the network. The batch size is set to 64, comprising of 16 identities with 4 images per identity. The Adam optimizer with a learning rate of $3.5\times 10^{-4}$ is used for optimization at the initial step and the learning rate of the backbone is set to $lr/10$ during the incremental learning. The model is trained for 40 epochs per step, and the early-stage learning regularization is performed during the first ten epochs. The hyper-parameter $T$, $\lambda_1$, and $\lambda_2$ are set to $0.5, 0.9$ and $0.6$  (see hyper-parameter analysis in the supplemental material).

\noindent{\textbf{General setup}} assumes that in most cases, there are more unseen classes initially emerging in a new camera than seen ones, and the number of unseen classes will increase linearly over time. As shown in Fig.~\ref{fig:3}(a), the identity distribution of the new data is managed by sampling the data from the initial camera. For example, we sampled 300, 500,and 300 identities in the first step for Market-1501, MSMT17 and DukeMTMC, respectively, yielding 110/131, 54/83, 74/144 seen/unseen classes in the second step. 

\noindent{\textbf{Exceptional setup}} is further considered for extreme scenarios where the majority of the classes captured by new cameras are old ones. This can be achieved by increasing the number of classes sampled in the initial step, thereby increasing the proportion of old classes in the new cameras.

For comparative experiments, we reproduce the state-of-the-art methods, i.e., data-free methods LwF~\cite{li2017learning}, AKA~\cite{pu2021lifelong}, AGD~\cite{lu2022augmented}, PatchKD~\cite{sun2022patch} , the replay-based methods iCaRL~\cite{rebuffi2017icarl}, PTKP~\cite{DBLP:conf/aaai/GeDWXYHZ22}, and a distribution alignment methods MMD~\cite{long2015learning} on our setting.  It is noteworthy that these methods are based on a class-disjoint setting, and they do not match our setting. Therefore, to implement them in our setting, they can only treat old classes as new ones. For more extensive assessment, we design some other comparative methods, including the baseline described in section~\ref{sec:baseline}, the fine-tune method that fine-tunes the model on new data, the Joint-T that denotes an upper-bound by training the model on all data seen so far.

\begin{table*}[t]
\caption{Ablation study of the contribution of ExtendOVA components during every incremental step in the  exceptional setup.}
\resizebox{.95\linewidth}{!}{
\begin{tabular}{cccc|c|cccc|cccc|cccc}
\toprule[1pt]
      &            &            &          
 & & \multicolumn{4}{c|}{Market-1501}                                                         &\multicolumn{4}{c|}{MSMT17}                                                & \multicolumn{4}{c}{DukeMTMC}  \\
           &            &            &            &   &       \multicolumn{2}{c}{Step2}       & \multicolumn{2}{c|}{Step3}  &       \multicolumn{2}{c}{Step2}   & \multicolumn{2}{c|}{Step3}&       \multicolumn{2}{c}{Step2}   & \multicolumn{2}{c}{Step3} \\ \cmidrule(lr){6-7} \cmidrule(lr){8-9}  \cmidrule(lr){10-11}  \cmidrule(lr){12-13}  \cmidrule(lr){14-15}  \cmidrule(lr){16-17}  
\multirow{-3}{*}{$\mathcal{L}_{\text{ID}^*}$} &
  \multirow{-3}{*}{$\mathcal{L}_\text{Aux}$} &
  \multirow{-3}{*}{$\mathcal{L}_\text{CD}$} & 
  \multirow{-3}{*}{EMA} &
  \multirow{-3}{*}{Method} &
  mAP &
  R-1 &
  mAP &
  R-1 &
  mAP &
  R-1 &
  mAP &
  R-1&
  mAP &
  R-1 &
  mAP &
  R-1  \\ \hline
      &         &      &       &      &71.3  & 85.0 & 62.7 & 83.3 & 37.4& 60.7 & 30.2 & 59.8&56.1 &70.5 & 51.3& 68.3\\
\Checkmark &    \Checkmark       &    &     & \multirow{-2}{*}{LwF} &         73.2  &    85.9     &     70.8  &85.3  &43.9 & 70.1 &38.4& 69.3 &61.5 &73.3 &54.6 &70.8  \\ \cline{5-17} 
\Checkmark & \Checkmark & \Checkmark &   \Checkmark &                    & 78.8 & 89.5 & 75.6   & 88.2& 50.0 &74.2 &46.3 &74.4 &64.1&76.5&62.4 &75.8\\
        
\Checkmark &            & \Checkmark &     \Checkmark &                       & 76.0                       & 87.7 & 73.9  & 87.4 & 49.3 & 73.6 & 45.4 & 73.3 &63.8 & 75.9 & 58.7&73.3\\

\Checkmark &
  \Checkmark &
   &\Checkmark & 
   &
  75.1&
  87.9 & 73.8
   &87.8 &48.5 & 73.7 & 44.4 & 72.7 &63.4 &75.2 &58.5 &71.1
   \\ 
\Checkmark &
  \Checkmark &
   \Checkmark &
   & \multirow{-3}{*}{ ExtendOVA} &
  74.9&
  86.3 & 72.6
   &86.8 & 45.7 & 72.9 & 44.2 & 73.1 &62.0 &74.0 &55.3 &71.1
   \\ \Checkmark &
  \Checkmark &
   &
   & &
  71.4&
  85.3 & 67.6
   &83.9 & 42.1 & 68.4 & 37.7 & 68.2 &58.1 &71.9 &51.9 &68.2
   \\  \bottomrule[1pt]
\end{tabular}}

\label{tab:ablation1}
\end{table*}

\subsection{Comparative Results with Different Settings}
We compare our ExtendOVA with the current state-of-the-art. The evaluation is conducted on all cameras encountered so far and the final-step results are reported in Tabel~\ref{tab:3} with both the general and exceptional setups. We summarize the results as follows:
\begin{itemize}
    \item Our proposed ExtendOVA outperforms the current state-of-the-arts by a clear margin, and is the closest to the upper bound Joint-T. Notably, it even achieves comparable or better results than the replay-based methods, validating the effectiveness of our proposed solutions.
    
    \item Previous methods, designing for the non-overlapping setting, still achieve poor performance. We attribute this poor performance to two aspects: Firstly, in the absence of cross-camera labels, these methods fail in learning cross-view representations. Secondly, turning old classes into new ones results in a false-positive prediction of the spurious classes, leading to accumulated errors.
    
    \item Interestingly, our proposed baseline outperforms the data-free methods LwF, AKA and AGD, demonstrating the potential improvement in addressing the class-overlap issue.
\end{itemize}

\subsection{Ablation Study}

\noindent{\textbf{A closer look at early stage regularization.}} In Fig.~\ref{fig:4}, we plot the per-sample probability distribution of confidence scores generated by different methods. As can be seen, the baseline method uses the maximum probability as the confidence score, resulting in significant confusion between seen and unseen classes. Higher threshold values will reject a large number of seen classes. While the output of the OVA detector is more discriminative, there is still significant noise introduced due to domain shift. Our method improves upon the OVA detector by incorporating early regularization learning, which significantly mitigates the noise caused by domain shift. 

\begin{figure}[t]
  \centering
  \includegraphics[width=\linewidth]{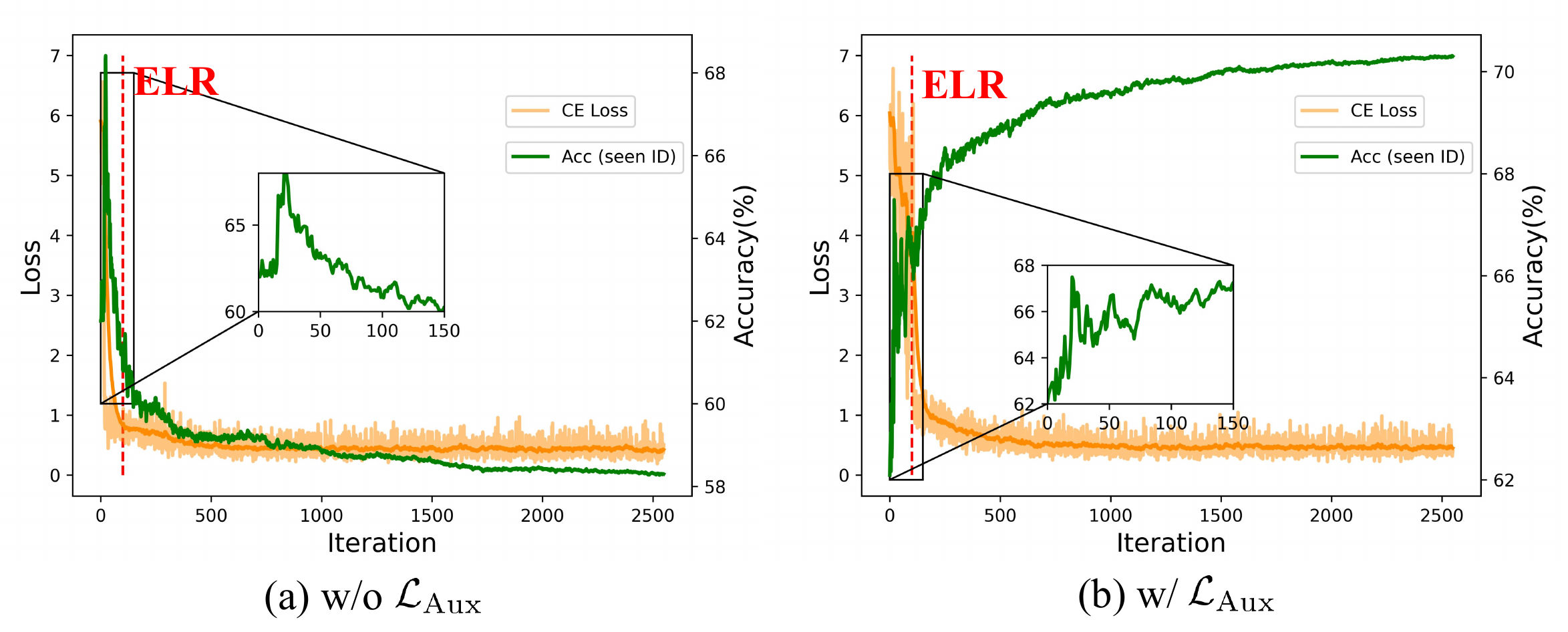}
  \caption{Curves of CE loss and model performance on seen classes. ELR: Early stage learning regularization. }
\label{fig:5}
\vspace{-0.2cm}
\end{figure}

To further observe the impact of early regularization learning on the ID-wise predictive performance, Fig.~\ref{fig:5} shows the training curves of ID loss and  model accuracy of seen classes during the training process. We can observe that during the training process, both models show a decreasing trend in loss and eventually converge. In the initial iterations of training, the accuracy of both models shows an increasing trend, indicating that the models have not yet started fitting to the noise. After a certain number of iterations, the accuracy of the model without early regularization starts to gradually decrease, while our method corrects the noise in the early stage, resulting in an increase in accuracy.
\begin{table}[t]
\centering
\caption{Compared with different distillation loss in the general setup. Base. is trained with $\mathcal{L}_{\text{ID}^*}+ \mathcal{L}_{\text{Aux}}$. }

\begin{tabular}{l|cc|cc}
\toprule[1pt]

 &
  \multicolumn{2}{c|}{Market-1501} &
  \multicolumn{2}{c}{DukeMTMC}  
\\

  \multirow{-2}{*}{Method} &
mAP &
  Rank-1 &
mAP &
  Rank-1 \\ \hline

  Base. & 57.8
 &  78.4 & 44.1
  & 62.7

 \\ 

  Base. $+\mathcal{L}_{\text{KD}}$ &
  62.5 &
  81.0 &
  47.5
&65.3
 \\

 Base. $+\mathcal{L}_{\text{CD}}$ &
   \textbf{65.5} &
 \textbf{83.4} &
\textbf{48.6} &
  \textbf{65.6}
 
 \\ \hline

  Base. $+$EMA &66.4
  &83.7
   &48.7
 &66.7
 
 \\

  Base. $+\text{EMA} +\mathcal{L}_{\text{KD}}$ &67.1
 &84.2
  &50.5
&68.4
 \\

 Base. $+\text{EMA} +\mathcal{L}_{\text{CD}}$ & \textbf{70.3}
  &\textbf{86.6}
  &\textbf{53.4}
 &\textbf{70.2}

  \\ \bottomrule[1pt]
\end{tabular}
\label{tab:3}
\end{table}

\textbf{Effectiveness of the different components.} We conduct ablation studies in the three-step exceptional setup to  evaluate the effectiveness of each module in each step. To evaluate the effectiveness of our ID-wise pseudo label generation module, we conduct experiments where we disable the $\mathcal{L}_\text{Aux}$ components and also  compare the results to a baseline method, i.e. LwF, which is trained using ReID loss supervised by intra-camera labels. First, as shown in Table~ \ref{tab:ablation1}, removing $\mathcal{L}_\text{Aux}$ will decrease the final performance by \bm{$0.9\%$} to \bm{$3.7\%$} in mAP.  Second, the combination of $\mathcal{L}_\text{Aux}$ and $\mathcal{L}_{\text{ID}^*}$ bring the gain in the range of \bm{$3.3\%$} to \bm{$8.2\%$} in mAP compared with the baseline. This suggests that simply optimizing the cross-entropy loss with intra-camera supervision is not sufficient. 

To evaluate the contribution of the EMA scheme, we conduct experiments by removing it.  Without the EMA technique, the performance drop ranges between \bm{$2.1\%$} and \bm{$4.3\%$} in the second step in terms of mAP, and the degradation becomes more significant as the incremental training phase proceeds. This clearly indicates that incorporating this design is crucial for overall performance. While the effect of $\mathcal{L}_\text{CD}$ is not as pronounced as the EMA scheme, it still has an obvious impact on the performance. When both terms are eliminated, there is a significant decline in performance, suggesting that the $\mathcal{L}_\text{CD}$ plays a role in maintaining the overall performance.

\begin{figure*}[ht]
  \includegraphics[width=\textwidth]{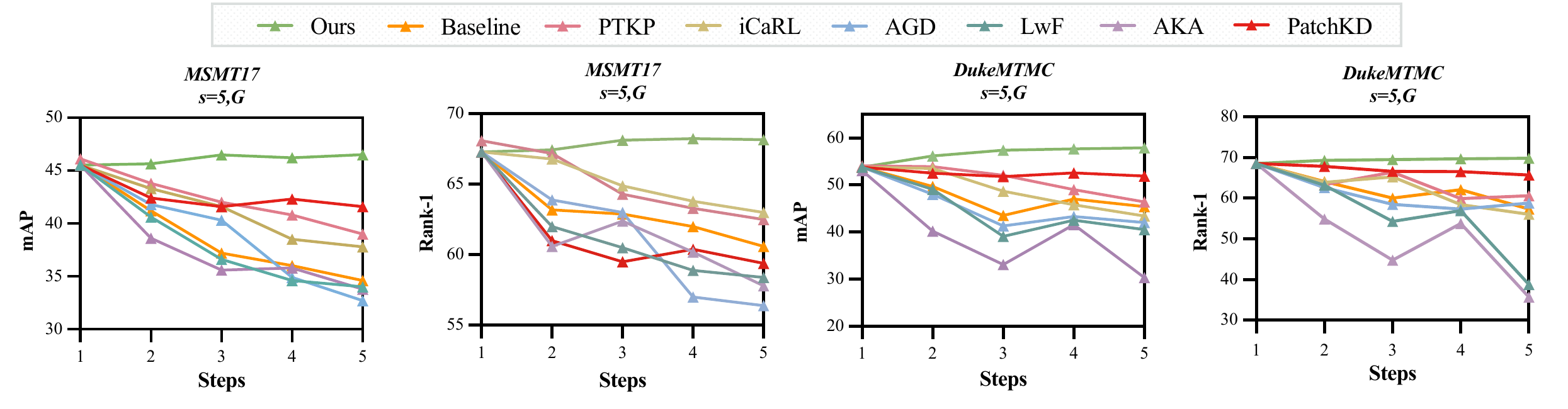}
  \centering
   \setlength{\abovecaptionskip}{-0.1cm}
  \caption{ Anti-forgetting evaluation on MSMT17 and DukeMTMC in the general setup. mAP and Rank-1 score on the test set of original cameras (test set on step 1) during the training process.  }
  \vspace{-0.2cm}
  \label{fig:6}
\end{figure*}

\textbf{Compared with different distillation loss.}
Table \ref{tab:3} compares the performance of our method with different distillation losses in the general setup. Specifically, we evaluate the impact of knowledge distillation (KD) and cross-camera distillation (CD) loss on the performance of the baseline model trained with ID loss ($\mathcal{L}_{\text{ID}^*}$) and the auxiliary losses ($\mathcal{L}_{\text{Aux}}$). From the table, we observe that incorporating distillation losses, either $\mathcal{L}_{KD}$ or $\mathcal{L}_{CD}$, improves the performance of the baseline model in terms of mAP and Rank-1 accuracy. Notably, adding $\mathcal{L}_{CD}$ achieves higher performance than adding $\mathcal{L}_{KD}$, indicating that $\mathcal{L}_{CD}$ is more effective in preserving the structure-wise knowledge.

\begin{table}[t]
\centering
\caption{Performance (\%) of seen classes identification by our proposed method. We report the scores obtained in the second step.}
\resizebox{\linewidth}{!}{
\begin{tabular}{c|c|cc|cc|cc}
\toprule[1pt]
&
 &
  \multicolumn{2}{c|}{Market-1501} &
  \multicolumn{2}{c|}{MSMT17}  &
  \multicolumn{2}{c}{DukeMTMC}\\

\multirow{-2}{*}{Setup} &
  \multirow{-2}{*}{Method} &
  \textit{Prec} &
  \textit{Recall} &
  \textit{Prec} &
  \textit{Recall} & \textit{Prec} &
  \textit{Recall}  \\ \hline
 &
  Baseline &
  98.4 &
  56.4& 97.5
  & 40.6& 51.7 & 40.5

 \\ 
 &
  OVA &
  93.3 &
  63.6 &
98.2 &
 64.3 & 98.3 & 60.9
 \\
\multirow{-3}{*}{\textbf{\emph{G}}} &
   ExtendOVA &
   89.8 &
 \textbf{80.0} &
 95.5 &
  \textbf{74.5} & 92.9 &\textbf{86.5}
 
 \\ \hline
 &
  Baseline &99.0
  &56.7
   & 99.2
 &33.5
 &94.4& 66.9

 \\
 &
 OVA &97.9
 &61.1
  &98.5
&40.1
   & 95.7 &70.5 \\
\multirow{-3}{*}{\textbf{\emph{E}}} &
ExtendOVA & 96.5
  &\textbf{75.6}
  &98.8
 &\textbf{51.8}
 & 95.0 & \textbf{77.9}
  \\ \bottomrule[1pt]
\end{tabular}}
\label{tab:4}
\end{table}

\begin{table}[t]
\centering
\caption{Results on the multiple-camera introduced setup. }

\begin{tabular}{c|cc|cc|cc}
\toprule[1pt]

 &
  \multicolumn{2}{c|}{Market-1501} &
  \multicolumn{2}{c|}{MSMT17}  
  & \multicolumn{2}{c}{DukeMTMC}  
\\

  \multirow{-2}{*}{Method} &
mAP &
  Rank-1 &
mAP &
  Rank-1 &
  mAP &
  Rank-1\\ \hline

 AKA~\cite{pu2021lifelong} & 57.9
 &  69.4 & 32.5
  &60.6& 45.2 &62.1

 \\ 
 AGD~\cite{lu2022augmented}   & 62.8
 &  72.0 & 31.6
  & 65.1 & 42.4 &60.6
 \\

 ExtendOVA &
  \textbf{70.6}  &
 \textbf{86.7} &
   \textbf{36.8}
  &\textbf{72.0}&
\textbf{52.5} &
  \textbf{68.9}

  \\ \bottomrule[1pt]
\end{tabular}
\label{tab:5}
\end{table}


 \subsection{Anti-Forgetting Evaluation}

We evaluate the anti-forgetting properties of our proposed method by measuring the performance on the test-set of the first step after each step. Fig.~\ref{fig:6} plots the forgetting trend on DukeMTMC and MSMT17 in the general setup. We found that our method showed superior anti-forgetting properties, with no performance degradation and even a slight improvement on the previous tasks. Our baseline model exhibits less forgetting when compared to data-free methods, clearly indicating that class-overlap is an issue to be addressed. AGD employs the DeepInversion~\cite{yin2020dreaming} to generate synthetic exemplars from previously learned classes, however, it still suffers from catastrophic forgetting. This is due to its reliance on a unified classifier, treating overlapping classes as new ones can result in the generation of a significant amount of noise.

\subsection{Further discussion}
\noindent{\textbf{Seen classes identification.}}
To further study the potential of identifying seen classes, we compare our method with the baseline and OVA. We use precision (prec) and recall as metrics. Precision is calculated as the percentage of truly seen classes among the selected classes, and recall is calculated as the percentage of selected seen classes among all seen classes in the new data. The results in Table \ref{tab:4}  show that our proposed method outperforms both the baseline and OVA methods in identifying seen classes, with comparable precision but higher recall scores in all datasets. This indicates that our method can effectively identify the seen classes in the new data, even when the data contains the domain shift. 

~\\
\noindent{\textbf{Extension to multiple introduced cameras.}}
To validate that our method can be applied in scenarios with multiple cameras increasing, we conducted further evaluations by including 2 cameras in the incremental step. Table~\ref{tab:5} presents the performance of the final training process in this setting. Notably, our method continues to outperform other state-of-the-art methods even when additional cameras are introduced. This finding suggests that our method is capable of addressing a universal CIPR problem.

\section{Conclusion}
\label{sec:6}
In this paper, we come up with a new yet very practical task, i.e., Camera Incremental person ReID (CIPR). We particularly emphasize the class-overlap issue brought by CIPR where the new camera might contain identities seen before and the ideal global cross-camera annotations are absent. To approach this task, we design a novel framework called ExtendOVA. In ExtendOVA, we address the class overlap issue by exploiting a One-vs-All detector combined with an early-stage regularization term to achieve the global
pseudo-label assignment. Extensive experiments verify the effectiveness and superiority of ExtendOVA.

\bibliographystyle{ACM-Reference-Format}
\bibliography{ref}

\end{document}